\documentclass[journal]{IEEEtran}
\ifCLASSINFOpdf
\else
\fi

\usepackage{graphicx}
\usepackage{multirow}
\usepackage{url}

\begin{document}
%
\title{Learning to Measure Changes: Fully Convolutional Siamese Metric Networks for Scene Change Detection}
%
%
%


\author{Enqiang~Guo,
        Xinsha~Fu,
        Jiawei~Zhu,
        Min~Deng,
        Yu~Liu,
        Qing~Zhu,
        and~Haifeng~Li\IEEEauthorrefmark{2}
\thanks{\IEEEauthorrefmark{2}Corresponding Author (email: lihaifeng@csu.edu.cn)}
\thanks{E. Guo and X. Fu are with the School of Civil Engineering and Transportation,
South China University of Technology, Guangzhou 510640, China (email: ctgmayday1997@mail.scut.edu.cn; fuxinsha@163.com).}
\thanks{J. Zhu, D. Min and H. Li are with the School of Geosciences and Info-Physics, Central South University, Changsha 410083,China}
\thanks{Y. Liu is with Institute of Remote Sensing and Geographic Information System, Peking University, Beijing, China}
\thanks{Q. Zhu is with  Faculty of Geosciences and Environmental Engineering, Southwest Jiaotong University, Chengdu 611756, China.}}

%
%

\markboth{IEEE TRANSACTIONS ON MULTIMEDIA,~Vol.~XX, No.~XX, XX~2018}%
{Shell \MakeLowercase{\textit{et al.}}: Bare Demo of IEEEtran.cls for IEEE Journals}
%



\maketitle

\begin{abstract}
A critical challenge problem of scene change detection is that noisy changes generated by varying illumination, shadows and camera viewpoint make variances of a scene difficult to define and measure since the noisy changes and semantic ones are entangled. Following the intuitive idea of detecting changes by directly comparing dissimilarities between a pair of features, we propose a novel fully \underline{Co}nvolutional \underline{si}amese \underline{m}etric \underline{Net}work(CosimNet) to measure changes by customizing implicit metrics. To learn more discriminative metrics, we utilize contrastive loss to reduce the distance between the unchanged feature pairs and to enlarge the distance between the changed feature pairs. Specifically, to address the issue of large viewpoint differences, we propose Thresholded Contrastive Loss (TCL) with a more tolerant strategy to punish noisy changes. We demonstrate the effectiveness of the proposed approach with experiments on three challenging datasets: CDnet, PCD2015, and VL-CMU-CD. Our approach is robust to lots of challenging conditions, such as illumination changes, large viewpoint difference caused by camera motion and zooming. In addition, we incorporate the distance metric into the segmentation framework and validate the effectiveness through visualization of change maps and feature distribution.The source code is available at \url{https://github.com/gmayday1997/ChangeDet}.
\end{abstract}


%
\IEEEpeerreviewmaketitle

\section{Introduction}
%
%
%
%

When a person is asked to determine the changes in a scene at different times$\{T_0,T_1\}$, it is natural to detect changes based on a pixelwise comparison between a pair of images, and then the changes in the scene can be inferred according to the degree of semantic dissimilarity. Recently, the state-of-the-art scene change detection (SCD) algorithm\cite{khan2016learning}\cite{alcantarilla2018street}\cite{sakurada2015change} are nearly based on a fully convolutional network (FCN), which are not intuitive because FCN-based models detect changes by learning a decision boundary with maximizing the margin rather than directly measuring the dissimilarities or changes.

Instead of simple cast change detection for classification, we strive to present a novel approach to detect changes by directly measuring the dissimilarities or changes between pairs of images at different times. The core value behind this intuition lies in regarding changes as dissimilarities. Moreover, to achieve that, a key question needs to be asked: how is a dissimilar function or metric to measure changes defined?

From the perspective of changes, it contains changes of interest, called semantic changes, and nuisance changes, called noisy changes. Given a pair of images, change detection aims to identify semantic changes at different times\cite{alcantarilla2018street}. However, the critical challenge in this task is noisy changes generated by challenging factors such as varying illumination, shadows and camera viewpoint differences that are difficult to distinguish from semantic changes, making changes difficult to define and measure owing to the noisy changes and semantic changes that are entangled. Intuitively, if one wants to explore the semantic changes and suppress the noisy changes, a feasible method is to learn more discriminative metrics to measure changes and assign semantic changes a higher measurement value, and noisy changes or no changes a lower measurement value.

\begin{figure}[t]
\centering\includegraphics[width=1\linewidth]{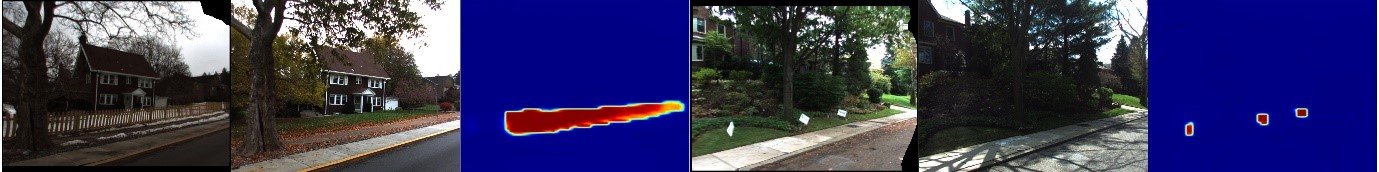}
\centering\includegraphics[width=1\linewidth]{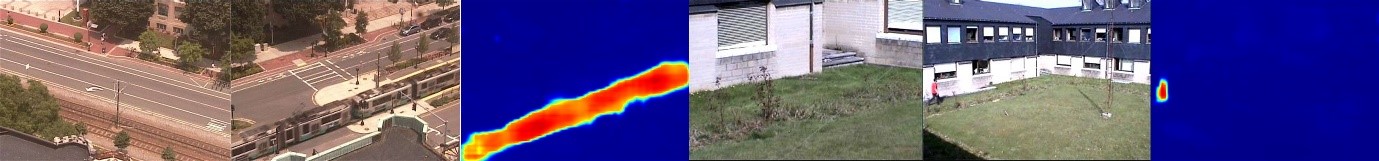}
\caption{Example visualization results of the change map by our proposed method under the challenging conditions of illumination changes, small viewpoint differences, and large viewpoint differences caused by camera rotation, and zooming.}
\end{figure}

As mentioned above, the solution is customizing a discriminative metric to distinguish the semantic changes from the noisy changes. However, it is difficult to explicitly obtain such a metric function. Recently, deep metric learning has been a key element in learning more discriminative features in computer vision tasks, such as face recognition\cite{sun2014deep}\cite{schroff2015facenet}\cite{deng2018arcface}\cite{wen2016discriminative} and feature learning\cite{huang2016learning}\cite{xing2003distance}\cite{sohn2016improved}. Its core idea is encouraging reducing intra-class variance and enlarging interclass differences. This learning strategy provides a feasible solution; that is, we can learn such an implicit metric by using a deep neural network as a universal approximation. Specifically, we define the changed area at the same position across the image pair as changed pairs, which are called positive-pairs in deep metric learning, and unchanged areas at the same position cross-image pair are called unchanged pairs, which are also named negative-pairs. From the view of deep metric learning, we try to learn an implicit metric subject to the distance of the unchanged-pairs being as small as possible, and the changed-pairs being as large as possible.

In our work, we propose a novel change detection framework, named the fully \underline{Co}nvolutional \underline{si}amese \underline{m}etric \underline{Net}work(CosimNet). Instead of simple classification, we leverage the intuitive idea of directly comparing a pair of images by customizing a discriminative implicit metric. It contains two parts: the deep features extracted from the Fully Convolutional Siamese Network (FCSN) and the predefined distance metric. All of the procedures can be seen as learning a dissimilar function directly on raw images. Our approach is robust to many challenging conditions, such as illumination changes and viewpoint differences. More examples are illustrated in Figure 1.

Our main contributions are as follows:

(1)	We propose a novel deep metric learning-based scene change detection that is able to directly measure changes using the learned implicit metric, which casts the change detection task to an implicit metric learning problem. To the best of our knowledge, this is the first time to propose a unified architecture to address lots of challenging conditions using an end-to-end deep metric learning method, especially in the case of large viewpoint of difference.

(2) We develop a Thresholded Contrastive Loss (TCL) to overcome noisy changes caused by large camera viewpoint differences, which presents a significant improvement. More details will be discussed in V.A

(3)	Compared with the baseline, the proposed approach achieves state-of-the-art performance on both the PCD2015 and VL-CMU-CD datasets and achieves competitive performance on the CDnet dataset.

(4)	Following the idea of learning a more discriminative feature, we integrate the distance metric into the baseline based on the FCN architecture, giving rise to better performance. We also find a reasonable explanation from the visualization of the change map.

\section{Related Works}\label{sec:Related Works}

Scene change detection (SCD) is a fundamental task in the field of computer vision \cite{khan2016learning}\cite{alcantarilla2018street}\cite{sakurada2015change}\cite{lee2000fast}\cite{st2015subsense}\cite{bianco2017combination}
\cite{braham2017semantic}\cite{lelescu2003statistical}.The core idea of SCD is detecting changes in multiple images $\{I_1,I_2,бн.I_M \}$ of the same scene taken at different times. The definition of changes can be divided into semantic changes and noisy changes\cite{alcantarilla2018street}. Specifically, semantic changes can be defined as changes in the scene caused by the disappearance or reconstruction of objects, such as building reconstruction and vehicle movement. Meanwhile, in terms of various challenging factors, noisy changes\cite{alcantarilla2018street} are divided into radiometric changes (illumination intensity variation, shadow, and seasonal changes) and geometric changes (viewpoint differences caused by camera rotation and zooming)\cite{radke2005image}. Obviously, we are more interested in semantic changes than noisy changes. However, noisy changes will definitely affect the appearance of an image, which leads to 'semantic fusion', especially under the conditions of a large viewpoint difference. How to correctly detect semantic changes and suppress noisy changes is still challenging in this task.

\begin{figure}[t]
\centering\includegraphics[width=1\linewidth]{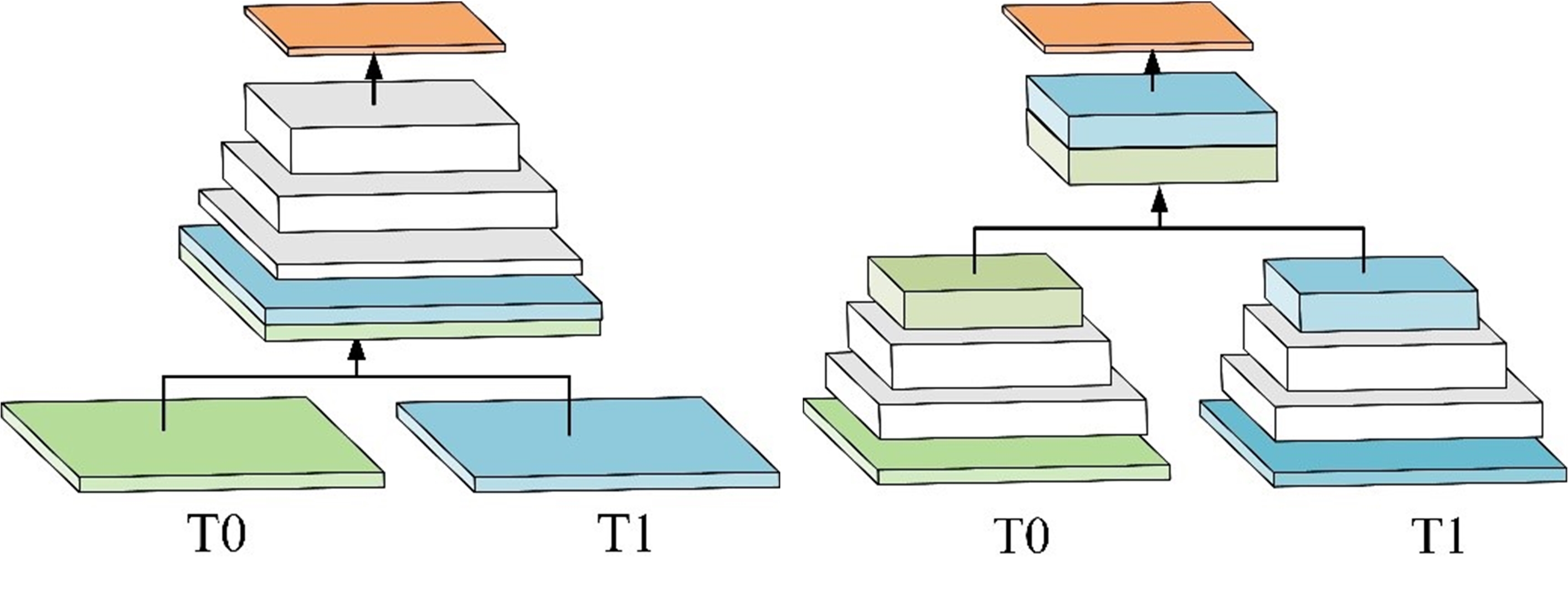}
\caption{Illustration of two change detection architectures based on FCN. The left one is an early-fusion architecture, and the right one is a late-fusion architecture.}
\end{figure}

The most traditional and classical scene change detection methods are the image difference method \cite{rosin1998thresholding}\cite{rosin2003evaluation}, which generates a change map by identifying the set of pixels that are 'significantly different' between two images and then obtaining a binary mask by thresholding. The advantage of this method is its small computational cost, while the drawback is that raw RGB features are incapable of effectively differentiating between semantic changes and noisy changes. To obtain more discriminative features, image rationing\cite{mahmoodzadeh2007digital}, change vector analysis\cite{bruzzone2002adaptive}\cite{bovolo2007theoretical}, Markov random field\cite{moser2011multiscale}, and dictionary learning\cite{mahmoodzadeh2007digital}\cite{lu2017joint} are proposed to address this issue. However, limited by the representation of hand-designed features\cite{radke2005image}, traditional methods are still sensitive to noisy changes, including illumination changes or viewpoint differences.

Recently, the convolutional neural network framework has achieved outstanding performance in computer vision tasks \cite{he2016deep}\cite{girshick2014rich}. In particular, almost all of the state-of-the-art change detection methods\cite{alcantarilla2018street}\cite{sakurada2015change} are based on fully convolutional networks (FCN)\cite{shelhamer2017fully} owing to the high precision in the dense-prediction task \cite{chen2018deeplab}\cite{xie2015holistically}\cite{dai2017deformable}\cite{kang2018depth}. Moreover, as shown in Figure 2, the SCD method based on FCN can be classified into early fusion\cite{zagoruyko2015learning}\cite{alcantarilla2018street} and late fusion, which both indicate detected changes by learning a decision boundary.However, learning a decision boundary still cannot answer three critical questions: (1) What are changes? (2) How can we measure changes? (3) Does there exist a suitable metric to measure changes that has a higher measurement value for changed pairs and a lower measurement for unchanged pairs?

To address these issues, we propose a novel approach to detecting changes which regards changes as semantic similarity by directly measuring changes with distance metrics, aiming to learn effective features that bring together unchanged pairs and separate changed pairs. Our work is based on the idea proposed in \cite{sakurada2015change}, which also utilized a distance metric to describe changes. However, the approach in \cite{sakurada2015change} used pretrained features extracted from the VGG model without fine-tuned learning, which are not sufficiently discriminative to describe changes. Instead, we propose an end-to-end trainable approach to learn discriminative features with a contrastive loss function, which customizes powerful features for the SCD task. The most relevant to our work is \cite{zhan2017change}, which also use contrastive loss to learn metrics. Beyond that,we propose a unified approach to address more challenging conditions, such as using thresholded contrastive loss to overcome large viewpoint differences.

\section{Proposed Approach}\label{sec:Approch}
\subsection{Overview}
\begin{figure*}[t]
\label{Arc}
\centering\includegraphics[width=1\linewidth]{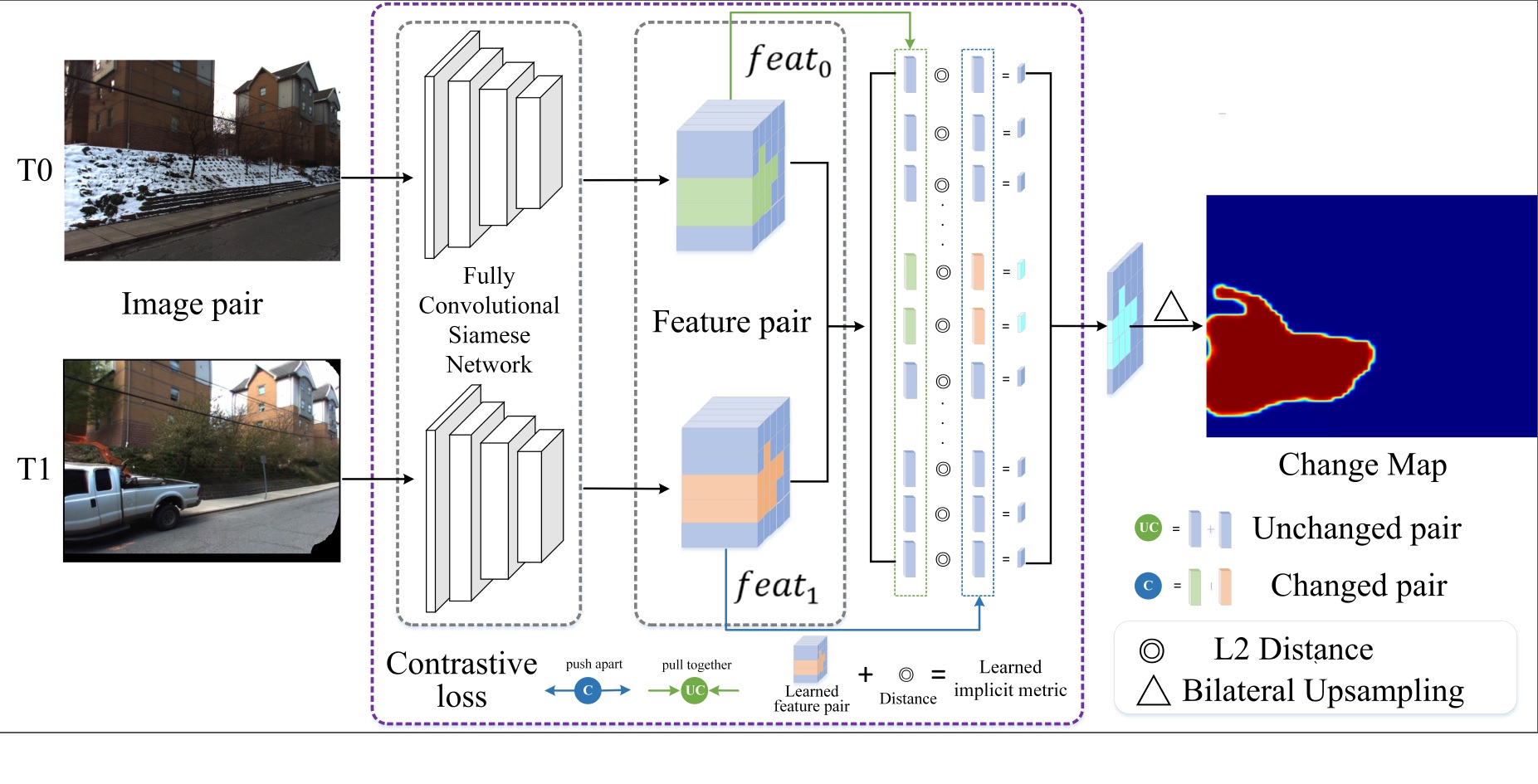}
\caption{ Illustration of the proposed architecture, named the fully \underline{Co}nvolutional \underline{s}iamese \underline{m}etric \underline{Net}work(CosimNet). Given a pair of images as input, we forward propagate the input through a full convolutional siamese network to generate feature pairs. Then, we utilize a simple predefined distance metric (l2 or cos) to measure the dissimilarity of the feature pairs. We named the unified processing including deep features and predefined distance metric as an implicit metric. To obtain a better implicit metric, we use the contrastive loss to bring together unchanged pairs and separate changed pairs. Finally, we use simple bilateral upsampling to the original spatial size.}
\end{figure*}

In this section, we will describe the proposed approach, named CosimNet, showing how to utilize a suitable metric to measure changes. As illustrated in Figure 3, a raw image pair $(X_0,X_1) \epsilon R^{3\times H\times W}$is first fed into a convolutional siamese network to generate a feature pair $(feat_0, feat_1)\epsilon R^{C \times h \times w}$. Then, we utilize a simple predefined distance metric whose value varies from 0 to 1, for instance, Euclidean distance (l2) or cosine similarity (cos), to produce a change map. The change map indicates how much confidence applies to the changes. We define the changed area at the same position across the image pair as changed-pairs and the unchanged area at the same position cross-image pair as unchanged pairs. As mentioned above, we named the unified processing including deep features and predefined distance metric as an implicit metric, which is used to measure change. Thus, the key lies in customizing an appropriate metric to obtain a higher distance value for changed pairs and a lower distance value for unchanged pairs. To achieve this goal, the contrastive loss was leveraged to pull together unchanged pairs and push apart changed pairs. In addition, a thresholded contrastive loss is also proposed to address the challenging issues of large viewpoint differences, which achieves significant improvement. In the following section, we will discuss more details about the implicit metric and metric learning with a loss function in III.B and III.C.

\subsection{Learning an Implicit Metric}

Following the basic idea of regarding changes as dissimilarity, we detect changes based on directly comparing and measuring the dissimilarity between pairs of images at different times. Clearly, the key factor lies in how to customize a suitable distance metric to measure the dissimilarities or changes in our work. In general, this dissimilar function contains two parts, feature descriptors, such as HOG\cite{dalal2005histograms} and a distance metric, for instance, Euclidean distance or Mahalanobis distance; however, it is still impaired by nuisance caused by viewpoint differences or illumination variances.

To address this limitation, we use a deep convolutional network, specifically, a siamese network, to learn more discriminative features. Siamese networks are widely used to address various visual tasks, such as patch matching \cite{zagoruyko2015learning}\cite{han2015matchnet}, flow estimation\cite{bailer2017cnn}, object tracking\cite{bertinetto2016fully}, and face recognition \cite{sun2014deep}\cite{schroff2015facenet}\cite{deng2018arcface}\cite{wen2016discriminative}. In detail, the siamese net contains two branches that share the same convolutional architecture and the same set of weights. The backbone of the siamese net can be any popular architecture, such as Googlenet\cite{szegedy2015going} or DeepLab\cite{Chen2014Semantic}. The fundamental capability of the siamese net is mapping pairs of images into pairs of features $(X_0,X_1) \epsilon R^{3\times H\times W}$. To measure the dissimilarity of feature-pairs, we first constrain this embedding to live on the C-dimensional hypersphere, i.e.,$\|feat\|=1$, and then, build a simple predefined metric over the normalized features.

The most popular predefined distance metrics are Euclidean distance\cite{harley2017segmentation}\cite{schroff2015facenet}\cite{wen2016discriminative}and cosine similarity\cite{sun2014deep}\cite{mueller2016siamese}. Selecting a suitable predefined distance metric heavily affects the performance of the model and depends on the corresponding task. For example, face recognition usually uses Euclidean distance, while cosine similarity is suitable for text processing tasks. In our work, we designed a comparative experiment over both distance metrics above and provided a quantitative analysis using contrast sensitive metrics, named RMS contrast. More details are described in V.B.

\subsection{Learning Discriminative Metric}
\subsubsection{Contrastive Loss}

As mentioned above, how to learn a discriminative metric to achieve outstanding performance in terms of this task, which gives a higher measurement value for a changed pair and a lower measurement value for an unchanged pair, is another core value in our work. Motivated by this idea, contrastive loss, aiming to enlarge the interclass difference and reduce the intraclass variation simultaneously, was adopted to supervise CosimNet to learn a good implicit metric. The contrastive loss was formulated as follows:

\begin{equation}
\label{Contrastive Loss}
Contrastive Loss=\left\{
\begin{array}{rcl}
D(f_i,f_j)       &      & {y_{i,j}=1}\\
max(0,m-D(f_i,f_j))     &      & {y_{i,j}=0}\\
\end{array} \right.
\end{equation}

where $f_i$ $f_j$ are feature vectors extracted from the feature pair at the same position, $D (f_i,f_j)$ measures the distance between $f_i$ and $f_j$ using Euclidean distance. $y_{(i,j)}=1$ indicates that there is no change at this location. In this case, the loss function tries to minimize the distance between $f_i$ and $f_j$. Whereas $y_{(i,j)}=0$  indicates that there is a change in this spatial position, encouraging the distance to be larger than the margin, denoted m. In addition, in terms of cosine similarity, we use the formulations as follows:

\begin{equation}
\label{Cosine}
CosLoss = \sum_{k=0}^{h\times w}{(y_k - e^{-\|w_k \times D_k(f_i,f_j)+ b_k\|})}^{2}
\end{equation}
where $D_k(f_i,f_j)$ is the cosine similarity between feature vectors. $w_k$ and $b_k$ are learnable scaling and shifting parameters.

\subsubsection{Thresholded Contrastive Loss}

Our goal is to robustly measure changes under any challenging outdoor conditions, especially in the case of large viewpoint differences caused by camera rotation or zooming. In that case, the original contrastive loss suffers drawbacks, such as poor performance and slow convergence during optimization. The main reason is the existence of the following two contradictions: On the one hand, a large viewpoint difference easily activates too much irrelevant information due to heavily unregistration, inevitably leading to 'semantic fusion' because the features of the unchanged pair and that of the changed pair are entangled; on the other hand, the original contrastive loss aiming to minimize the distance to zero of the feature pair extracted from the region that contains a large visual difference, which contributed to zero prediction to have a relatively good performance.

The critical issue that leads to this contradiction is that it is unreasonable to make the distance of the semantic dissimilar feature-pair equal to 0. To address this limitation, we attempt to adopt a more flexible strategy to optimize noisy change. We slightly modify the contrast loss by setting a margin denoted as ${\tau}_{k}$, which is referred to as TCL, indicating that it is not necessary to minimize the distance to zero. A similar idea was also proposed in \cite{bailer2017cnn}, and the specific formulation is shown in equation 3.

\begin{equation}
\label{ThresholdedContrastiveLoss}
TCL=\left\{
\begin{array}{rcl}
D(f_i,f_j)-\tau_{k}       &      & {y_{i,j}=1}\\
max(0,m-D(f_i,f_j))     &      & {y_{i,j}=0}\\
\end{array} \right.
\end{equation}

To demonstrate the outstanding performance with TCL under the condition of a large viewpoint difference, we performed numerous comparative experiments with different thresholds on the CD2014 dataset. More details are described in V.B.

\subsection{Training Policy}

To enhance more discriminativeness for the SCD task, the MultiLayer Side-Output (MLSO) training policy was adopted in our work, which was first proposed in deeply supervised nets\cite{lee2015deeply}\cite{xie2015holistically}. The MLSO was designed based on the following two observations. (1) In the training phase, the supervised information of single-layer loss gradually decreased with the layerwise backward propagation, leading to less discriminative features of intermediate layers when supervised information was lost. (2) In the test phase, the representation of the upper layer feature heavily depends on the discriminativeness of the intermediate features.

Inspired by this, we introduced a companion loss function, specifically, contrastive loss, to supervise feature learning of the intermediate layer, which can be seen as an additional constraint on the upper hidden layers. During the training process, to balance the loss between the different layers, we introduce layer-balancing weights, termed $\beta_h$. The specific formula is as follows:

\begin{equation}
\label{layer_wise}
Loss = \sum_{l=h}^{L}{\beta_h \times loss_h}
\end{equation}

where $loss_h$ denotes the loss between the distance map and the ground truth. In the inference phase, we also set different confidence thresholds with respect to different layers, and the final prediction is the average of all outputs.

\begin{figure}[t]
\centering\includegraphics[width=1\linewidth]{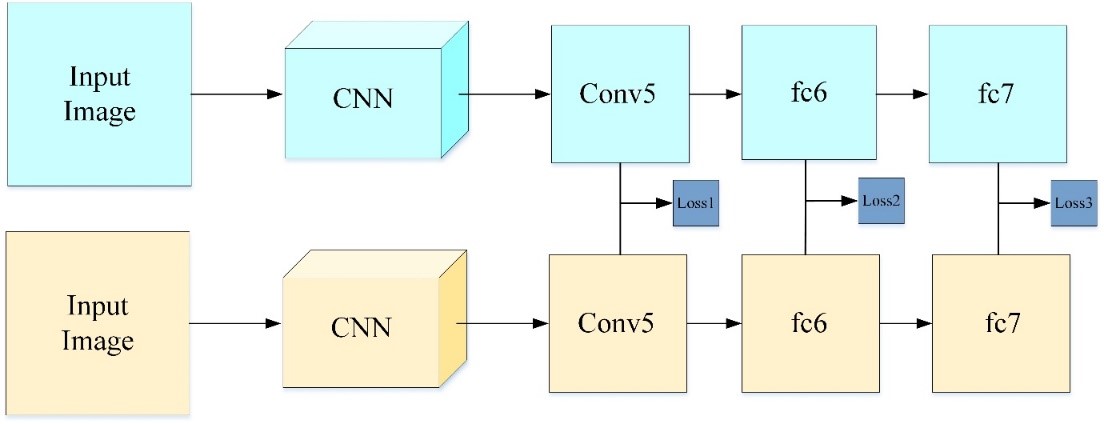}
\caption{Illustration of the multilayer side output training policy}
\end{figure}


\section{Experiment and Discussions}

In this section, we describe our experimental evaluation and provide an ablation study of our proposed architecture. We show competitive performance compared to baselines on the CD2014, VL-CMU-CD and PCD2015 datasets.

\subsection{Implementation Details}

In the experiment, the backbone of CosimNet was based on DeeplabV2\cite{chen2018deeplab}, whose last classification layer was removed. During fine-tuning, the learning rate of the top five layers was initialized as 1e-7, and fc6 and fc7 were set to 1e-8. We trained all the models using a stochastic gradient descent algorithm with the momentum equal to 0.90 and the weight decay equal to 5e-5. All of the experiments in this paper were tested on the Pytorch\cite{paszke2017automatic}  platform, and the training hardware was a GTX TITAN 1080. In next section, we will report the performance of CosimNet on three datasets.

\subsection{ Datasets and Results}
\subsubsection{VL-CMU-CD Dataset}

The VL-CMU-CD \footnote{VL-CMU-CD Dataset Link: \url{https://ghsi.github.io/proj/RSS2016.html}}\cite{alcantarilla2018street} is a change detection dataset with a long time span and challenging changes, including semantic changes, such as new buildings and construction areas. and noisy changes, such as viewpoint changes, lighting conditions/weather/season changes. The dataset contains a total of 151 sequences, which equals 1362 image pairs and provides labeling masks for 5 classes, including vehicle and traffic signal. According to the data splits provided in \cite{alcantarilla2018street}, the splits contain a training set and a test set, with 97 sequences, which are 933 image-pairs in total, and 54 sequences, totaling 429 for each. During the training processing, we resize all of the samples to 512 x512 and convert a multi-class labeling mask to a binary change map, meaning that we only focus on changes instead of class information. The comparison of performance between the proposed approach and the state-of-the-art methods are shown in Table 1.

\begin{table}[!htbp]
\centering
\caption{Comparison of performance (F-Score) with the popular method over the VL-CMU-CD dataset. }
\begin{tabular}{c|c c|c c c|c }
\hline
\multirow{2}{*}{Method}& \multicolumn{2}{c|}{Distance} & \multicolumn{3}{c|}{Different layers} & \multirow{2}{*}{$F-Score$}\\
\cline{2-6} & $l2$ & $cos$ & $C_5$ & $F_6$ & $F_7$ & {}\\
\hline
Depth & {} & {} & {} & {} & {} & 0.22 \\
\hline
Dense SIFT\cite{kim2015dasc} & {}  & {} & {} & {} & {} & 0.24\\
\hline
DAISY & {} & {} & {} & {} & {} & 0.18 \\
\hline
DASC\cite{tola2010daisy} & {} & {} & {} & {} & {} & 0.23 \\
\hline
Sakurada\cite{sakurada2015change}  & {}  & {} & {} & {} & {} & 0.40\\
\hline
CDNet\cite{alcantarilla2018street} & {} & {} & {} & {} & {} & 0.55\\
\hline
\hline
CosimNet-$1$layer-$cos$ & {} & $\surd$ & {} & {} & $\surd $ & 0.638\\
\hline
CosimNet-$2$layer-$cos$ & {} & $\surd$ & {} & $\surd$ & $\surd $ & 0.644\\
\hline
CosimNet-$3$layer-$cos$ & ${} $ & $\surd$ & $\surd $ & $\surd $ & $\surd $ & 0.647 \\
\hline
CosimNet-$1$layer-$l2$ & $\surd $ & {} & {} & {} & $\surd $ & 0.678\\
\hline
CosimNet-$2$layer-$l2$ & $\surd $ & {} & {} & $\surd $ & $\surd $ & 0.695 \\
\hline
CosimNet-$3$layer-$l2$ & $\surd $ & {} & $\surd $ & $\surd $ & $\surd $ & 0.706\\
\hline
\end{tabular}
\end{table}

In terms of different settings, including various distance metrics (Euclidean distance, cosine similarity) and different training policies (MLSO), we designed a set of comparative experiments. As shown in Table 1, we compare CosimNet with other state-of-art models, showing significant improvement. Specifically, the CosimNet-3-layer-l2 achieves a $15\%$improvement, and even the CosimNet-1 layer-cos also has an $8\%$ improvement. In addition, we observe two phenomena according to the performance with different settings: (1) The MLSO training policy indeed improves the semantic representation of the middle layer. A powerful representation of the middle layer contributes to the improvement of performance. (2) In general, Euclidean distance outperforms the cosine similarity in measuring changes. To explore more insights in different metrics, visualization analysis of change maps and quantitative analysis using contrast sensitivity will be described in V.B.

\subsubsection{PCD2015 Dataset}

The PCD2015 dataset \footnote{PCD2015 Dataset Link: \url{http://www.vision.is.tohoku.ac.jp/us/research/4d_city_modeling/pano_cd_dataset/}} \cite{sakurada2015change}  contains two subsets, named Tsunami and GSV. In detail, Tsunami consists of 100 panoramic image pairs of scenes after a tsunami, and the GSV dataset contains 92 panoramic image pairs of Google Street View. In our experimental setting, we directly kept the original size of 1024x224 for training and performed 5-fold cross-validation at a ratio of 8:2 \cite{alcantarilla2018street}\cite{sakurada2015change}. Similar to the settings in IV.B, we also set 6 comparative experiments in terms of different factors. As shown in Table 2, we observe that CosimNet-3 layer-l2 has a $3\%$ improvement in the tsunami dataset and a nearly $8\%$ improvement in the GSV dataset.

\begin{table}[!htbp]
\centering
\caption{Comparison of performance (F-Score) with the baseline method over the PCD dataset}
\scalebox{0.9}{
\begin{tabular}{c|c c|c c c|c|c}
\hline
\multirow{2}{*}{Method}& \multicolumn{2}{c|}{Distance} & \multicolumn{3}{c|}{Different layers} & \multicolumn{2}{c}{$F-Score$}\\
\cline{2-8} & $l2$ & $cos$ & $C_5$ & $F_6$ & $F_7$ & Tsunami & GSV\\
\hline
Dense SIFT\cite{kim2015dasc} & {}  & {} & {} & {} & {} & 0.649& 0.528\\
\hline
Sakurada\cite{sakurada2015change}  & {}  & {} & {} & {} & {} & 0.724 & 0.639\\
\hline
CDNet\cite{alcantarilla2018street} & {} & {} & {} & {} & {} & 0.774 & 0.614\\
\hline
\hline
CosimNet-$1$layer-$cos$ & {} & $\surd$ & {} & {} & $\surd $ & 0.601 & 0.582\\
\hline
CosimNet-$2$layer-$cos$ & {} & $\surd$ & {} & $\surd$ & $\surd $ & 0.715 & 0.624\\
\hline
CosimNet-$3$layer-$cos$ & ${} $ & $\surd$ & $\surd $ & $\surd $ & $\surd $ & 0.745 & 0.672\\
\hline
CosimNet-$1$layer-$l2$ & $\surd $ & {} & {} & {} & $\surd $ & 0.776 & 0.674\\
\hline
CosimNet-$2$layer-$l2$ & $\surd $ & {} & {} & $\surd $ & $\surd $ & 0.784 & 0.688\\
\hline
CosimNet-$3$layer-$l2$ & $\surd $ & {} & $\surd $ & $\surd $ & $\surd $ & 0.806 & 0.692\\
\hline

\end{tabular}}
\end{table}

\subsubsection{CDnet Dataset}

The CDnet\footnote{CDnet Dataset Link:\url{http://www.changedetection.net/}}\cite{goyette2012changedetection}\cite{wang2014cdnet} dataset consists of 31 videos depicting indoor and outdoor scenes with boats, trucks, and pedestrians that have been captured in different scenarios. It contains a range of challenging factors, including dynamic backgrounds, camera jitter, shadow, intern object motion, PTZ, and night video, which aims to solve foreground detection in complex outdoor conditions. In general, foreground detection can be regarded as change detection based on multiframe sequences\cite{radke2005image}, which usually use image differencing as a common practice\cite{rosin1998thresholding}\cite{rosin2003evaluation}.

As shown in Figure 5, we selected the background images (i.e., without any foreground objects) as the reference image at time $T_0$and others as the query images at $T_1$. In detail, we built a total of 91595 image pairs, which consist of a training set and a validation set with 73276 pairs and 18319 for each. All images were scaled to $512 \times 512$ during training. As a prior, we directly compare ConsimNet-3-layer-l2, which performs best in the previous experiments with the state-of-the-art. The result \footnote{Evaluation on CDnet:\url{http://jacarini.dinf.usherbrooke.ca/results2014/516/}} comparison between the proposed method and other popular baselines is shown in Table 3.

\begin{figure}[t]
\centering\includegraphics[width=1\linewidth]{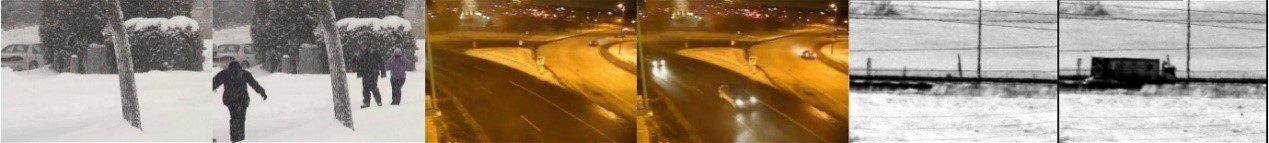}
\centering\includegraphics[width=1\linewidth]{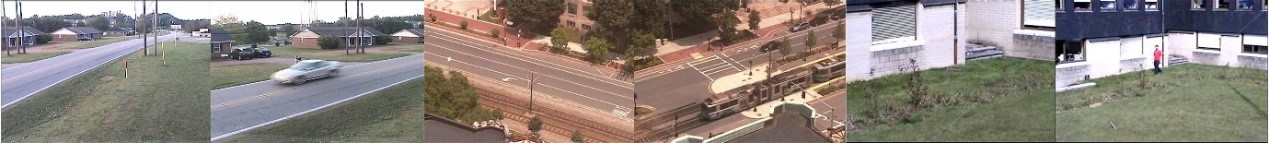}
\caption{ Challenging examples of image pairs at different times taken from the CDnet dataset. The first row shows registered images, and the second row shows the unregistered samples due to large camera viewpoint differences caused by camera motion or zooming.}
\end{figure}

\begin{table*}[tbp]
\centering
\caption{Result Comparison of foreground detection performance with other popular methods over CDnet dataset}
\begin{tabular}{c|c|c|c|c}
\hline
\textbf{Method} & \textbf{$Average FPR$} & \textbf{$Average FNR$} & \textbf{$F-Measure$} &\textbf{$Precision$}\\
\hline
\hline
\textbf{SuBSENSE\cite{st2015subsense}} & {0.0096} & {0.1876} & {0.7408} & {0.7509}\\
\textbf{IUTIS-3\cite{bianco2017combination}} & {0.0060} & {0.2221} & {0.7551} & {0.7875}\\
\textbf{SemanticBGS\cite{braham2017semantic}} & {0.0039} & {0.2625} & {0.7892} & {0.8305}\\
\textbf{CP3-online\cite{liang2015co}} & {0.0106} & {0.1771} & {0.7917} & {0.7663}\\
\textbf{Cascade CNN\cite{wang2017interactive}} & {0.0032} & {0.0494} & {0.9209} & {0.8997}\\
\hline
\hline
\textbf{CosimNet-$3$layer-$l2$}& {0.0007} & {0.1964}& {0.8591} & {0.9383}\\
\hline
\end{tabular}
\end{table*}

Compared with other state-of-the-art approaches, our model achieved competitive performance but was still insufficient in some metrics. The reason can be divided into two folds. On the one hand, an almost state-of-the-art approach utilizes semantic segmentation to remedy this task because of pixel-level annotations for each frame, which are free from challenging factors, such as large viewpoint differences. On the other hand, the proposed method is based on image differencing, whose performance is severely dependent on the background selection and image pair registration. Specifically, our model has presented significant improvement under the condition of large camera viewpoint differences due to some designs such as TCL; however, it still has lower precision in comparison with the semantic segmentation approach. More demo videos of change detection over CDnet can be found at \url{https://www.youtube.com/watch?v=VcJIpf_X-iA},\url{https://www.youtube.com/watch?v=trhQE4Uq-GM}

\section{Discussion}

To further demonstrate the effectiveness of CosimNet, we will discuss three challenging issues:
(1)	Is the CosimNet model robust to camera viewpoint differences caused by camera rotating and zooming?
(2)	The CosimNet framework can be regarded as an image differencing method in which a fixed threshold is required to obtain a binary mask. Thus, a natural question is the model performance sensitive to the threshold?
(3)	Can the idea of metric learning enforced by contrastive loss truly contribute to learning more discriminative features for change detection tasks?
To address these issues, we conducted experiments on the CD2014 and VL-CMU-CD datasets.

\subsection{Is It Robust to Viewpoint Difference?}

Distinguishing semantic changes from noisy changes is a key property of SCD tasks. The method proposed in this paper, following the idea of image differencing, is naturally sensitive to camera viewpoint differences, which are mainly caused by camera rotation or zooming. To address this challenging problem, it is a common practice to use preprocessing algorithms, such as SfM \cite{taneja2011image}, to align image pairs. However, this method is not only computationally expensive but also has limited effects. Thus, it is natural to ask the following question: is there a strategy to directly generate semantic changes from unregistered image pairs, suppressing noisy changes without any preprocessing or postprocessing methods? Considering that the performance of the model is subject to varying degrees of influence by different viewpoint differences, we evaluate the unified approach to address small viewpoint differences and large viewpoint differences with experiments on the VL-CMU-CD dataset.

(1)	Small Viewpoint Differences

    It is a common practice to give rise to small viewpoint differences because these images are taken at different times and hard to capture from the same viewpoint, especially in the case of a long time span. To address this problem, we directly try the original contrastive loss, formulated as equation 2, to measure changes under small viewpoint differences. As shown in Figure 6, we observe that a small viewpoint difference is insufficient to diminish the performance of CosimNet. The intuitive explanation is that the receptive field of the CNN is so large and gradually increases with layer depth, so that a small viewpoint difference is insufficient to affect the image appearance. Thus, an original contrastive loss can handle this challenging condition.

\begin{figure}[t]
\centering\includegraphics[width=1\linewidth]{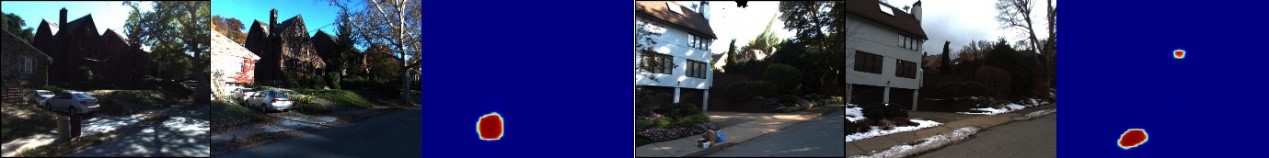}
\centering\includegraphics[width=1\linewidth]{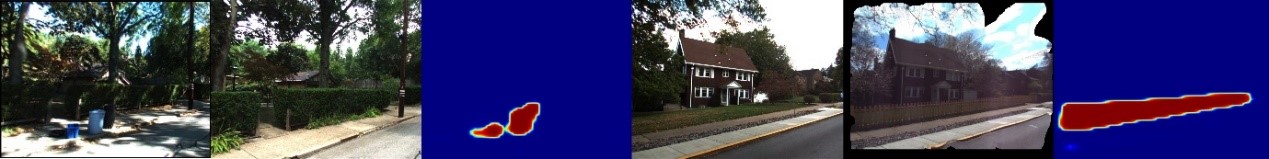}
\caption{The visualization results produced by CosimNet with an original contrastive loss under the condition of a small camera viewpoint difference}
\end{figure}

\begin{figure}[t]
\centering\includegraphics[width=1\linewidth]{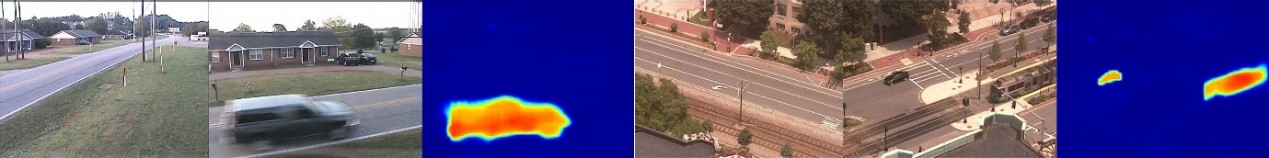}
\centering\includegraphics[width=1\linewidth]{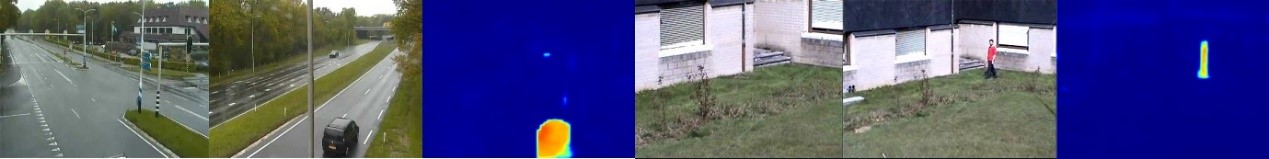}
\caption{The visualization results produced by CosimNet with TCL under the condition of a large camera viewpoint difference}
\end{figure}

(2)	Large Viewpoint Difference

  Among all of the noisy changes, a large viewpoint difference is the most challenging factor, which heavily reduces the performance of CosimNet. As already mentioned, we propose thresholded contrastive loss (TCL) to address this challenging problem. To evaluate the effectiveness of the proposed TCL loss, which was formulated as equation 3, we designed a set of comparative experiments over the cd2014 dataset, especially on the PTZ categories (continuous Pan, Intermittent Pan, TwoPositionPTZCam, and zoomInZoomOut). In terms of different thresholds, we set [0, 0.4] at an interval of 0.1, especially TCL, which is equivalent to the original contrastive loss when the threshold is zero. We explore the performance over different threshold settings.

  From the comparison results shown in Figure 8, we observe that:

  (1) in view of addressing the issue of large viewpoint differences, TCL clearly outperforms the original contrastive loss, meaning that more tolerance optimization is robust to these noisy changes;

  (2) CosimNet with TCL achieves the best performance at the threshold of 0.1. However, performance decreases when the threshold value is set to 0.4, suggesting that an overly tolerant training strategy may reduce inter-class differences. To further confirm the effectiveness of TCL, we visualize the change maps on the cd2014 dataset under a large viewpoint difference. As shown in Figure 7, despite being heavily unregistered between image pairs, we observe that the semantic change area always has the strongest response, while the noisy change area has a lower or zero response. In other words, ConsimNet with TCL indeed overcomes this challenging limitation, automatically ignoring noisy changes.

\begin{figure}[t]
\centering\includegraphics[width=1\linewidth]{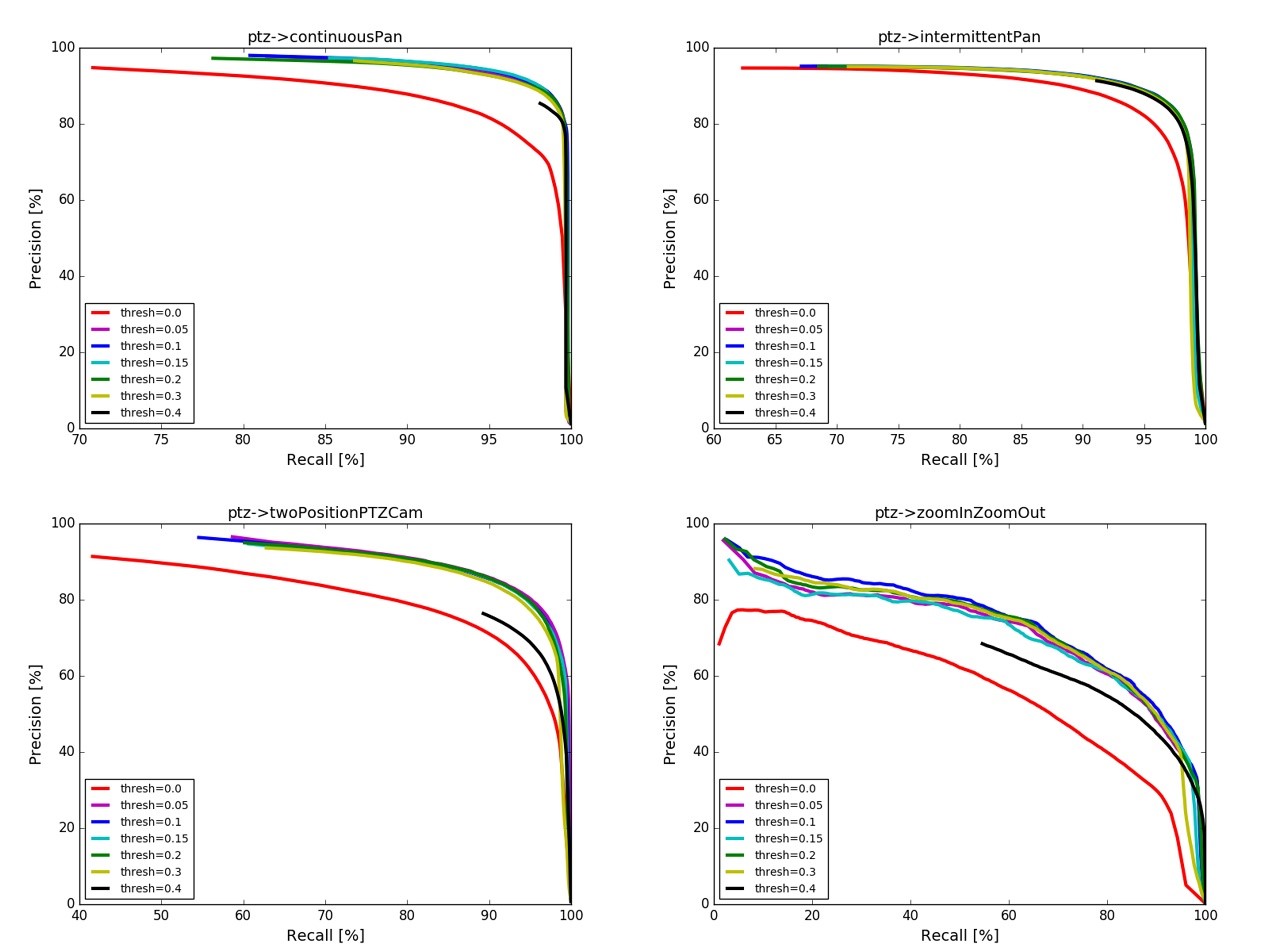}
\caption{ Precision-recall curve with different thresholds on the PTZ class of the cd2014 dataset; TCL is equivalent to original contrastive loss when the threshold is equal to zero.}
\end{figure}

\subsection{Contrast Sensitivity}

Considering that threshold selection can heavily affect the performance of our method, we utilize quantitative metrics, named threshold contrast \cite{ginsburg1983comparison}\cite{peli1990contrast}, to measure the contrast sensitivity, which defines the threshold between the foreground and the background in natural images. Specifically, there are many definitions of threshold contrast to measure the contrast sensitivity, such as Michelson contrast and RMS contrast\cite{peli1990contrast}. Among them, Michelson Contrast $C_M$  was suitable for repeating patterns such as sine waves, while RMS contrast was used for complex patterns, such as random dot patterns or natural images, which will be leveraged in our paper.


\begin{equation}
\label{Michelson}
C_M = \frac {L_{max}-L_{min}}{L_{max} + L_{min}}
\end{equation}

\begin{equation}
\label{RMS}
C_{RMS} = \frac {\sqrt{\frac{1}{N} \sum ({L_i -L_{mean}})^{2}}}{L_{mean}}
\end{equation}
In our work, we hope to maximize the contrast between the background and the foreground, which indicates changes so that the performance of the model will not heavily depend on the threshold selection. To confirm the effectiveness of our design choices, we analyze threshold contrast in two different design settings, including different distance metrics and features at the difference 'level', which also means features extracted from different layers.

\begin{figure}[t]
\centering\includegraphics[width=1\linewidth]{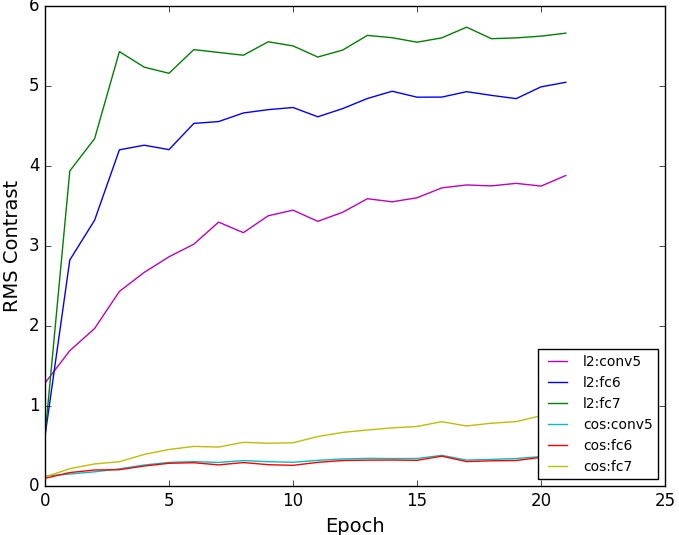}
\caption{RMS contrast with different distances (l2 and cosine) and layers (conv5, fc6 and fc7). There were 22 epochs in the x-axis, and RMS contrast values are shown in the y-axis}
\end{figure}

(1)	Different Distance Metrics

It is undoubted that different distance metrics have different abilities to measure the distance of feature pairs. From the perspective of the SCD task, the key principle of selecting a suitable metric is obtaining a higher distance value for the changed pair and a lower distance value for the unchanged pair. In the above experiment, we observe that performance with the l2 distance always outperforms that with cosine similarity. To explore the reasons, a qualitative analysis of two distance metrics with RMS contrast is illustrated in Figure 9.

We observe that the RMS contrast metric has a significant increase during the training process. According to the features at the same level, we found that the l2 distance has larger contrast values than cosine similarity, indicating that the l2 distance has a more powerful ability to distinguish changes from the background.

Moreover, to explore more insight in comparison with those two distant metrics, we visualize a change map to show which regions of change obtain the strongest response. As illustrated in Figure 10, the change map produced by the l2 distance highlights more relevant semantic changes and suppresses more noisy changes than cosine similarity for each layer, which demonstrates that the l2 distance outperforms cosine similarity in separating changed pairs and bringing together unchanged pairs.

(2) Semantic Feature at Different Level

It is well known that the features of deeper layers have richer semantic information than those of shallower layers \cite{zeiler2014visualizing}. Similar to the above analysis of Figure 9, according to the same distance metric, the RMS contrast value of the fc7 layer is larger than that of fc6 and conv5. Moreover, as illustrated in Figure 10, the change map at fc7 has a stronger response than that of the other two layers and indeed focuses on all the relevant semantic changes, meaning that more discriminative features lead to a more robust performance.

\begin{figure}[t]
\centering\includegraphics[width=1\linewidth]{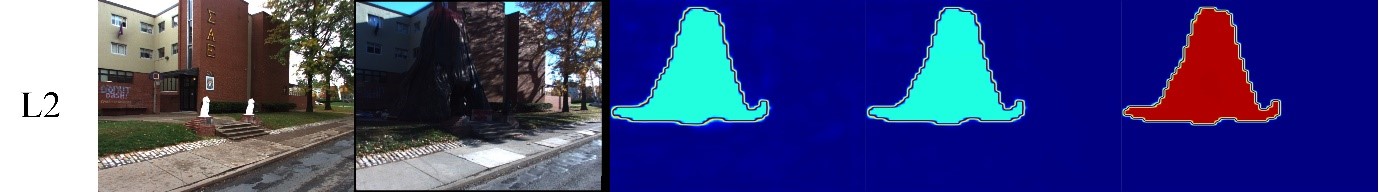}
\centering\includegraphics[width=1\linewidth]{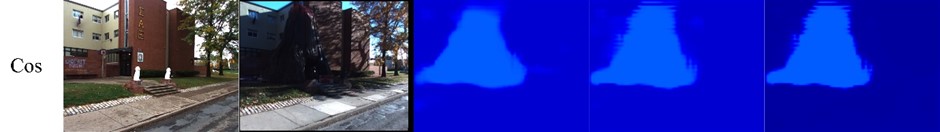}
\centering\includegraphics[width=1\linewidth]{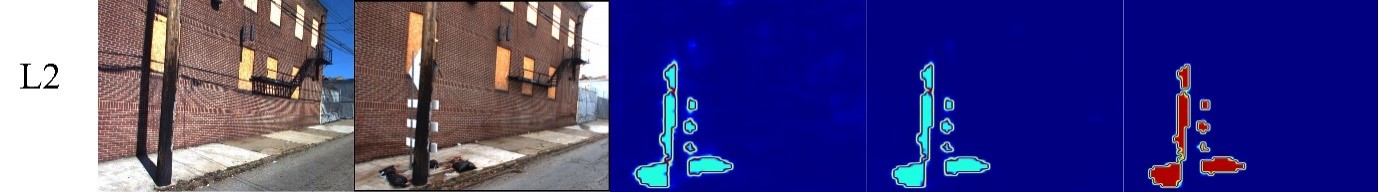}
\centering\includegraphics[width=1\linewidth]{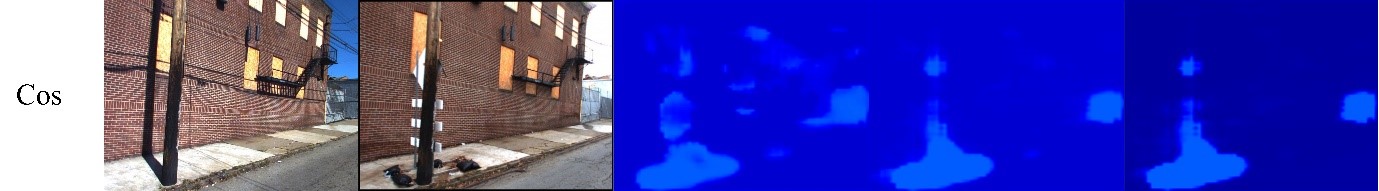}
\centering\includegraphics[width=1\linewidth]{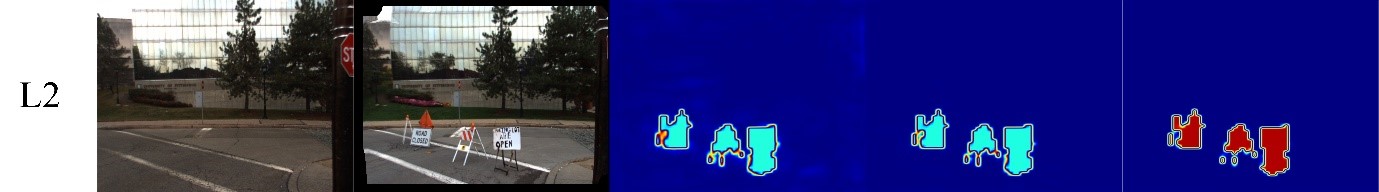}
\centering\includegraphics[width=1\linewidth]{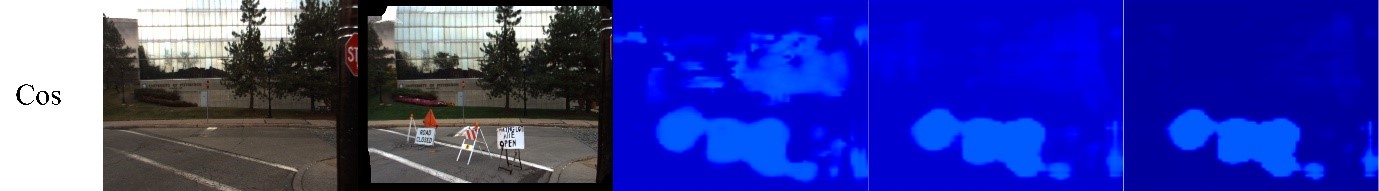}
\caption{Comparison of the change maps for different layers with different metrics, including l2 distance and cosine similarity. From left to right in each row, there are images at t0 time, images at t1 time, and change maps of conv5, fc6 and fc7, respectively (best viewed in color).}
\end{figure}

\subsection{Powerful Feature Representation in FCN Architecture}
\subsubsection{Learning Discriminative Feature}
How to improve the semantic discriminability of features is the core problem of computer vision tasks \cite{harley2017segmentation}\cite{schroff2015facenet}\cite{deng2018arcface}\cite{wen2016discriminative}. The key to solving this issue lies in increasing interclass differences and reducing intraclass variations. The FCN-based change detection method concatenates the dimensions of the feature-pair $\{feat_0^k,feat_1^k\}$ captured at different times as $ feat^k=concat(feat_k^0,feat_k^1)$. The framework essentially learns the decision boundaries between the different feature categories so that the category of a certain feature can be determined by the distance between the feature and the decision boundaries. Therefore, learning a discriminative feature is critical to the classification task.

The CosimNet introduces prior knowledge that can measure the change, giving a higher distance measurement value for a change pair and lower distance measurement for an unchanged pair. Similarly, in view of semantic feature learning, we also hope that the features $\{feat_0^k,feat_1^k\}$ are clustered together in the feature space, which precisely reflects the essential requirements of the discriminative learning of feature semantics.

Following the idea of improving discriminability of features, we incorporate distance metric learning into the model based on FCN, and use similarity learning to constrain features. In our implementation, we optimizing the model, based on DeeplabV2\cite{chen2018deeplab}, with multi-task loss function, formulated as equation 7. To be special, loss consists of two parts, where $Loss_{class}$ is cross entropy, used for pixel-level classification and $Loss_{feat}$ is contrastive loss, used for learning discriminative feature. In additional, in order to prevent the gradient domain during training, we set a constant $\lambda$ to offset the imbalance between two losses. In our experiment, we set $\lambda$ to 3.

Following the idea of improving the discriminability of features, we incorporate distance metric learning into the model based on FCN and use similarity learning to constrain the features. In our implementation, we optimized the model based on DeeplabV2 \cite{chen2018deeplab} with a multi-task loss function, formulated as equation 4. Specifically, loss consists of two parts, where $Loss_{class}$ is cross-entropy, used for pixel-level classification and $Loss_{feat}$ is a contrastive loss, used for learning discriminative features. In addition, to prevent the gradient domain during training, we set a constant $\lambda$ to offset the imbalance between two losses. In our experiment, we set $\lambda$ to 3.

\begin{equation}
\label{Multi-Task}
Loss = Loss_{class} + {\lambda} \times Loss_{feat}
\end{equation}

\begin{table}[tbp]
\centering
\caption{Comparison of performance (F-score) on three datasets}
\begin{tabular}{c|c|c|c}
\hline
\textbf{Method} & \textbf{$Tsunami$} & \textbf{$GSV$} & \textbf{VL-CMU-CD} \\
\hline
CosimNet-$3$layer-$l2$ & {0.806} & {0.692} & {0.706} \\
\hline
FCN-Later-Fusion & {0.809} & {0.685} & {0.714} \\
\hline
FCN-Metrics & {0.814} & {0.692} & {0.721} \\
\hline
\end{tabular}
\end{table}

The comparison results are shown in Table 4. Integration with deep metric learning achieves a minor improvement performance over the three datasets. To explore the main reason for the improvement, we provide a visual comparison between the original FCN and the distance metric learning. As shown in the third column of each row in Figure 11, the original FCN has limitations of keeping the object boundary smooth, for instance, bench (the first row) and traffic signal (the fourth and fifth row) were fragmented or mislabeled. From the perspective of classification, we can regard this as misclassification, but to be further explored in the field of feature learning, mislabeling was caused by intra-class inconsistency, which is also one of the drawbacks of FCN. Similarly, as shown in the last two columns of each row in Figure 11, we found that the model under the guidance of the change maps has significant enhancement with strong consistency in comparison with the original FCN. Specifically, integrating distance metrics can be regarded as additional constraints of feature learning, which reduces intra-class variance and contributes to smoothing boundaries.

\begin{figure}[t]
\centering\includegraphics[width=1\linewidth]{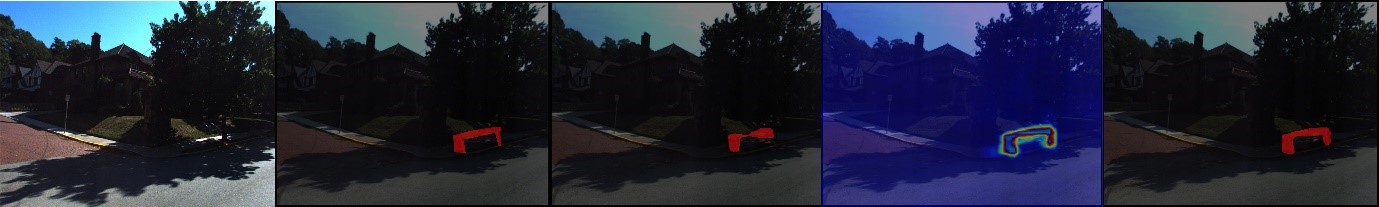}
\centering\includegraphics[width=1\linewidth]{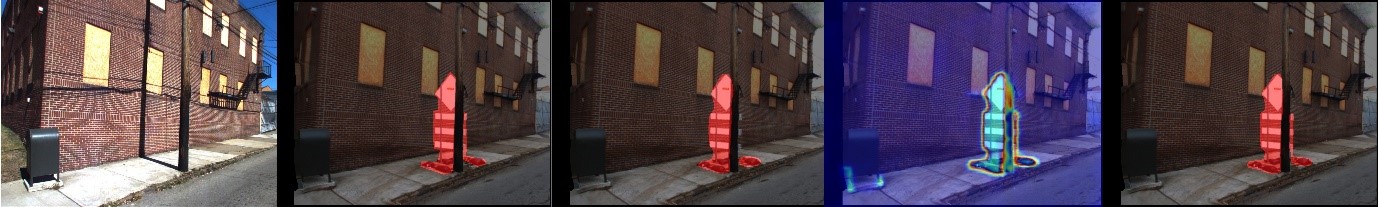}
\centering\includegraphics[width=1\linewidth]{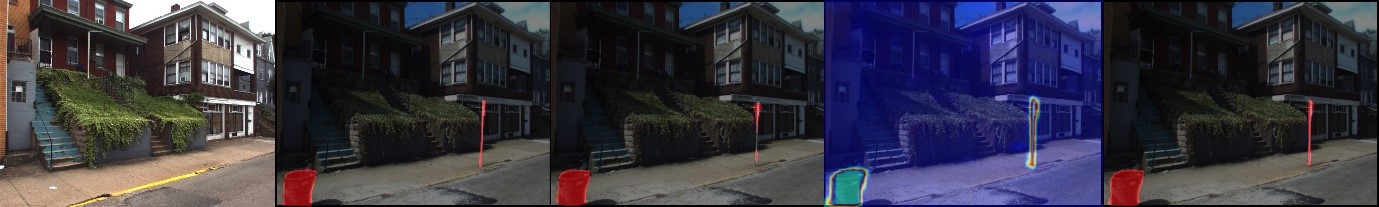}
\centering\includegraphics[width=1\linewidth]{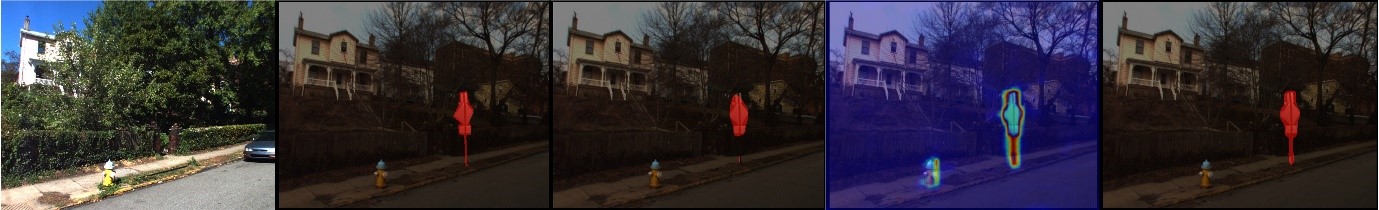}
\centering\includegraphics[width=1\linewidth]{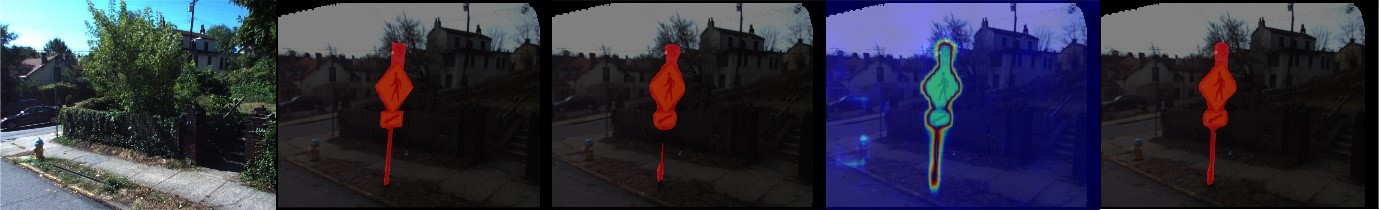}
\centering\includegraphics[width=1\linewidth]{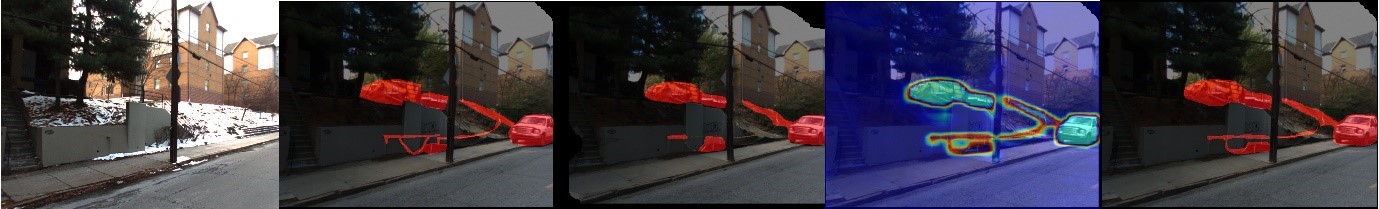}
\caption{From left to right in each row, there are images at $t_0$ time, images at $t_1$ time, the ground truth, late fusion prediction, change maps produced by our method, and predictions produced by our method. Best viewed in color.}
\end{figure}

\subsubsection{Feature Visualization}

To further explore the insights of enhancement by deep metric learning, we used the t-SNE\cite{maaten2008visualizing} algorithm to visualize the distribution of features in the fc7 layer of the FCN model and that of CosimNet. The results of the two-dimensional fc7 features are plotted in Figure 12 for illustration. From the comparison of the change detection results, the advantage of the proposed CosimNet is that it keeps the boundary smooth and makes objects consistent. Moreover, observing the corresponding two-dimensional feature distribution of these two approaches, we found that learned features from CosimNet are more separable and have smaller intra-class variations. Clearly, integrating into deep metric learning is indeed forcing intra-class compactness and interclass separability, which contributes to learning more discriminative or disentangled features, leading to better performance.

\begin{figure}[t]
\centering\includegraphics[width=1\linewidth]{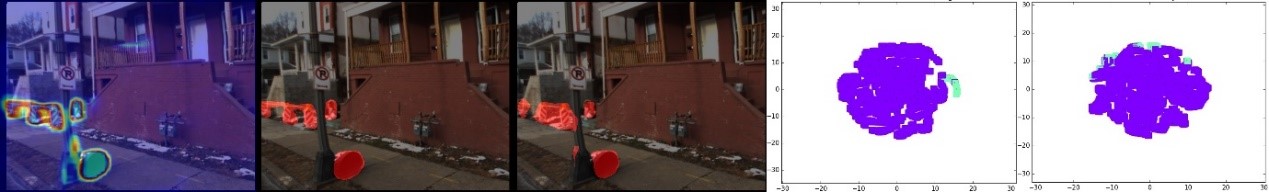}
\centering\includegraphics[width=1\linewidth]{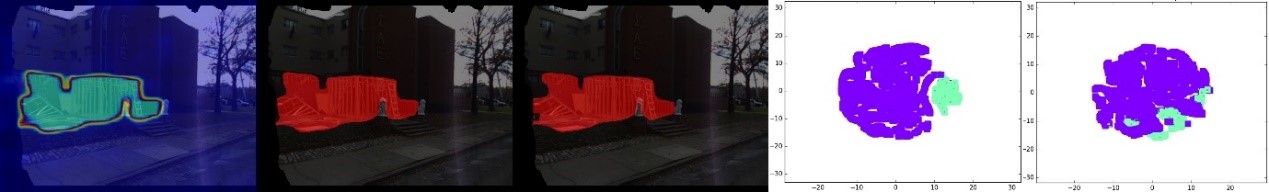}
\centering\includegraphics[width=1\linewidth]{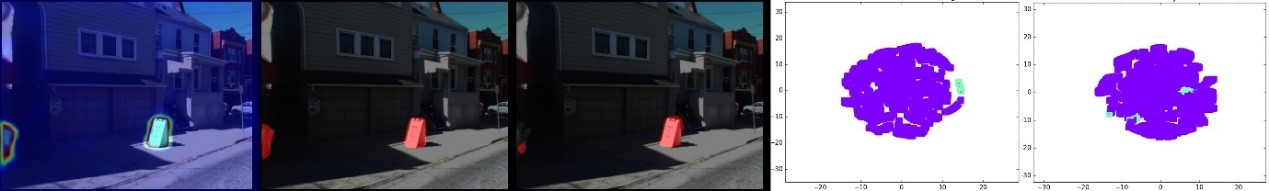}
\centering\includegraphics[width=1\linewidth]{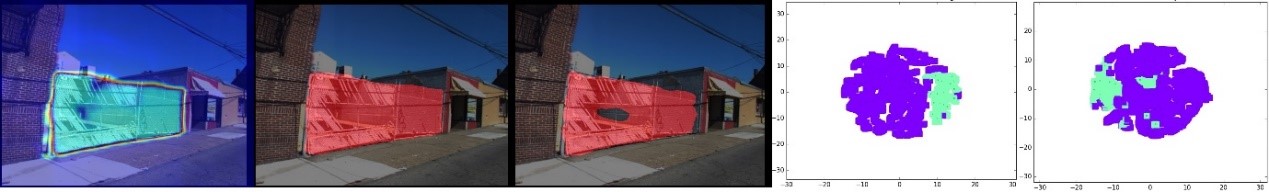}
\caption{From left to right in each row: (1) change map (2) prediction produced by CosimNet, (3) prediction produced by FCN, (4) two-dimensional fc7 feature embedding of CosimNet (5) two-dimensional fc7 feature embedding of FCN. As shown in the last two columns of each row, different colors denote different classes, where cyan means change-pair feature embedding and pure means unchanged one. (Fig is best viewed in color.)}
\end{figure}

\section{Conclusion}

We propose a novel framework, named CosimNet for the scene change detection task and measure changes directly using a learned implicit metric. To reduce the distance between the unchanged pair and increase the distance between the changed pair, this paper uses a siamese network for extracting features of image pairs and the contrastive loss to learn a better implicit metric. Specifically, we find thresholded contrastive loss with a more tolerant strategy to punish this noisy change which can address the issue of large viewpoint differences. Experiments on three popular datasets demonstrate the proposed method are robust to many challenging conditions, such as illumination variations, seasonal variations, camera motion, and zooming. The feature visualization in low dimension space illuminates that the CosimNet learns disentangled embeddings which distinguish change features and unchanged ones. The learned disentangled embeddings, which are considered as the very promising feature to a machine learning model, are the key fact to yield better performance of our model. Our framework potentially gives a new perspective to rethink how to measure "changes" in a SCD.

\appendices

\ifCLASSOPTIONcaptionsoff
  \newpage
\fi



%

\bibliographystyle{IEEEtran}
\bibliography{IEEEabrv,refer}

%




\end{document}